\documentclass[10pt,journal,compsoc]{IEEEtran}
\usepackage{booktabs}
\usepackage{multirow}
\usepackage{bm}
\usepackage{array}
\usepackage{graphicx}
\usepackage{epstopdf}
\usepackage{subfigure}
\usepackage{amssymb}
\usepackage{amsmath}
\usepackage{makecell}
\usepackage{bm}
\def\A{{\bf A}}
\def\B{{\bf B}}

\def\F{{\bf F}}

\def\S{{\bf S}}
\def\T{{\bf T}}

\def\v{{\bf v}}

\def\0{{\bf 0}}
\def\1{{\bf 1}}

\def\etal{{\em et al.}}
\def\eg{{\em e.g.}}
\def\ie{{\em i.e.}}

\def\etal{{\em et al.\/}\,}

\usepackage[pagebackref=false,breaklinks=true,letterpaper=true,colorlinks,bookmarks=false]{hyperref}
\usepackage{cite}
\usepackage[linesnumbered,ruled,vlined]{algorithm2e}
\usepackage{multirow}
\usepackage{color}
\usepackage{CJK}
\graphicspath{{Figures/}}

\begin{document}


\title{Inverse-Consistent Deep Networks for Unsupervised Deformable Image Registration}

\author{Jun~Zhang
		
\thanks{J.~Zhang is with the Tencent AI Medical Center, China. (Email: xdzhangjun@gmail.com).}

	
	
}

\IEEEtitleabstractindextext{	
\begin{abstract}
Deformable image registration is a fundamental task in medical image analysis, aiming to establish a dense and non-linear correspondence between a pair of images.
Previous deep-learning studies usually employ supervised neural networks to directly learn the spatial transformation from one image to another, requiring task-specific ground-truth registration for model training. Due to the difficulty in collecting precise ground-truth registration, implementation of these supervised methods is practically challenging. Although several unsupervised networks have been recently developed, these methods usually ignore the inherent inverse-consistent property (essential for diffeomorphic mapping) of transformations between a pair of images. Also, existing approaches usually encourage the to-be-estimated transformation to be locally smooth via a smoothness constraint only, which could not completely avoid folding (typically means registration errors) in the resulting transformation. To this end, we propose an \emph{I}nverse-\emph{C}onsistent deep \emph{Net}work (\emph{ICNet}) for unsupervised deformable image registration. Specifically, we develop an {\emph{inverse-consistent constraint}} to encourage that a pair of images are symmetrically deformed toward one another, until both warped images are matched. Besides using the conventional smoothness constraint, we also propose an {\emph{anti-folding constraint}} to further avoid folding in the transformation. The proposed method does not require any supervision information, while encouraging the diffeomoprhic property of the transformation via the proposed inverse-consistent and anti-folding constraints. We evaluate our method on T1-weighted brain magnetic resonance imaging (MRI) scans for tissue segmentation and anatomical landmark detection, with results demonstrating the superior performance of our ICNet over several state-of-the-art approaches for deformable image registration. Our code will be made publicly available.
\end{abstract}

}
\maketitle

\section{Introduction}
\label{Sec:introduction}
As a fundamental task in medical image analysis, deformable image registration aims to establish dense, non-linear spatial correspondences between a pair of images (denoted as the source/moving image and the target/fixed image)~\cite{sotiras2013deformable}. For example, while it is difficult to compare brain magnetic resonance imaging (MRI) scans of different subjects due to significant anatomical variability~\cite{liu2018landmark,lian2018multi}, deformable registration enables direct comparison of anatomical structures across scans, and thus, is  crucial for understanding variability across populations and the longitudinal evolution of brain anatomy for individuals with brain diseases~\cite{hajnal2001medical,avants2008symmetric,shen2002hammer}.

Over the past few decades, a variety of non-learning-based deformable registration methods have been proposed for medical image analysis~\cite{bajcsy1989multiresolution,shen2002hammer,rueckert1999nonrigid,thirion1998image}. Typically, these methods iteratively optimize a similarity function for each pair of images to non-linearly align voxels with the similar appearance, while encouraging local smoothness on the registration mapping~\cite{sotiras2013deformable}. Since the similarity function needs to be \emph{optimized from scratch} for each pair of unseen images, \ie, the inherent registration patterns shared across different images are ignored, these methods are usually slow to perform deformable registration in practical applications.
To address these issues, supervised learning-based and deep-learning-based approaches have been developed for deformable image registration~\cite{yang2016fast,cao2018region,cao2018deformable}. These methods typically rely on {\emph{task-specific ground-truth registration}} to train a regression model for image registration. However, the difficulty of collecting ground-truth information often limits their utility in real-world applications.
In addition, the mapping learned by these supervised methods might be biased by the selected ground-truth registration.

\begin{figure}[t]
	\setlength{\belowdisplayskip}{0pt}
	\setlength{\abovedisplayskip}{0pt}
	\setlength{\abovecaptionskip}{-2pt}
	\centering
	\includegraphics[width=0.46\textwidth]{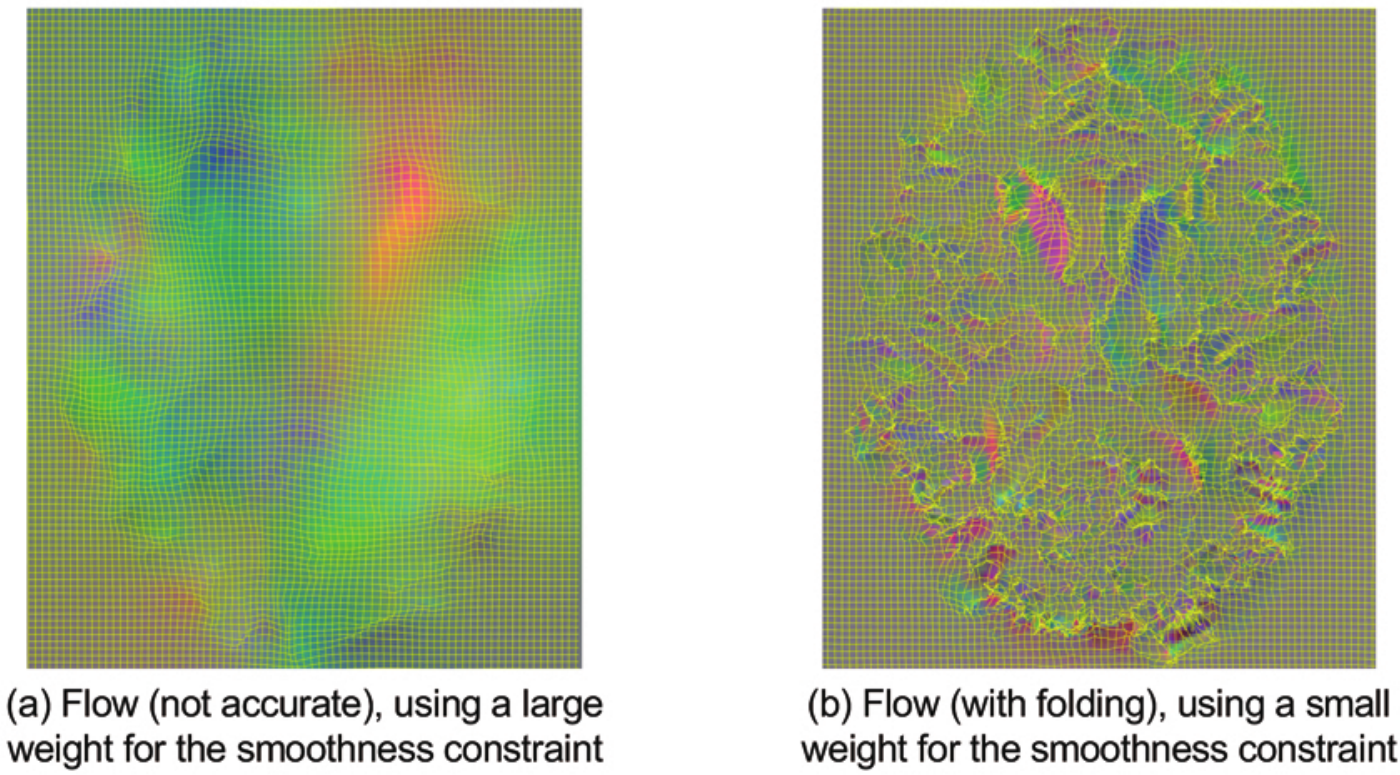}
	\caption{Illustration of two flows learned by a state-of-the-art deep-learning method~\cite{balakrishnan2018unsupervised} using (a) a large weight for the smoothness constraint and (b) a small weight for the smoothness constraint, respectively.}
	\label{fig_smootEffect}
\end{figure}

Recently, several unsupervised deep-learning methods have been proposed for medical image registration~\cite{wu2013unsupervised,miao2016cnn,shan2017unsupervised,de2017end,balakrishnan2018unsupervised} without using any pre-defined supervision information for network training, thus maintaining the unsupervised nature of deformable registration. Although these methods achieve better registration performance in comparison to traditional supervised-learning-based methods, the output transformations (\eg, displacement field or flow) are usually asymmetric, \ie, the inherent inverse-consistent property of transformations between a pair of images is ignored. Here, the inverse-consistent property means that to-be-learned optimal transformations would encourage that a pair of images are symmetrically deformed toward each another, and the two bidirectionally deformed images are finally matched. Unfortunately, previous studies usually independently estimate the transformation from an image $\A$ to an image $\B$ or from $\B$ to $\A$, thus failing to ensure these transformations be inverse mappings for each another. Besides, most of the existing (supervised or unsupervised) algorithms utilize solely a spatial smoothness penalty to constrain the transformation, which could not completely avoid foldings (typically indicating errors) in the registration mapping. As an illustration, in Fig.~\ref{fig_smootEffect}, we show two flows generated by a state-of-the-art deep-learning method~\cite{balakrishnan2018unsupervised} trained with different (\ie, strong vs. weak) contributions from the smoothness constraint.
As shown in Fig.~\ref{fig_smootEffect}~(a), if we excessively encourage local smoothness of the to-be-estimated flow by using a large weight for the smoothness constraint, the obtained registration will be inaccurate due to global errors. Otherwise, there will be a lot of foldings in the learned flow (see Fig.~\ref{fig_smootEffect}~(b)), thus generating wrong registration due to local defects. That is, it is challenging to properly tune the contribution of the smoothness constraint to simultaneously avoid foldings in the estimated flow and maintain high registration accuracy.

To address these two issues, in this paper, we propose an Inverse-Consistent deep neural Network ({\bf{ICNet}}) for unsupervised deformable image registration. Specifically, in ICNet, we develop an {\emph{inverse-consistent constraint}} to encourage that a pair of images are symmetrically deformed toward each another in multiple passes, until the bidirectionally deformed images are matched to achieve correct registration. Besides using the conventional smoothness constraint, we also develop an {\emph{anti-folding constraint}} to avoid foldings in the to-be-estimated flow. The proposed ICNet method does not require any supervision information, which also encourages the diffeomoprhic property of the transformations via the proposed inverse-consistent and anti-folding constraints.
We evaluate our method in both tasks of tissue segmentation and anatomical landmark detection with 3D T1-weighted brain MRI scans. The experimental results demonstrate the superior performance of the proposed method over several state-of-the-art methods in deformable image registration.

The rest of this paper is organized as follows. We briefly introduce relevant studies in Section~\ref{S2_relatedWork}. In Section~\ref{S3_method}, we present the proposed network, inverse-consistent constraint, and anti-folding constraint in detail. We then describe studied materials, competing methods, experimental settings and results in Section~\ref{S4_experiments}. We further analyze the influence of several essential strategies used in the proposed method in Section~\ref{S5_discussion}. We finally conclude this paper in Section~\ref{S6_conclusion}.

\section{Related Work}
\label{S2_relatedWork}
\subsection{Deformable Image Registration}
Deformable image registration refers to an non-linear process of revealing the voxel-wise spatial correspondence between source and target images. Let $\S$ denotes the source image, and $\T$ represents the target image. We assume $\F_{ST}$ is the to-be-learned flow (\ie, displacement field) that warps $\S$ to $\T$. The optimization problem is typically defined as
\begin{equation}
\small
\begin{aligned}
\mathcal{F}(\S,\T) = \mathcal{L}_{similar}{\left(\T, {\Phi}(\S,\F_{ST})\right)} + \mathcal{R}(\F_{ST}),
\end{aligned}
\label{eq_optimize}
\end{equation}
where ${\Phi}$ denotes the transformation operator that warps the source image $\S$ to the target image $\T$ using the flow $\F_{ST}$. The first term in Eq.~\ref{eq_optimize} is a similarity/matching/distance criterion, which is used to quantify the level of alignment between the warped source image ${\Phi}(\S,\F_{ST})$ and the target image $\T$. The second term is a regularizer that constrains the transformation to favor a specific property, such as encouraging the estimated flow to be locally smooth. The optimization problem consists of either maximizing or minimizing the objective function, depending on how the first term is defined.

Different algorithms for deformable registration mainly differ in deformable models, similarity criteria, and numerical optimization~\cite{Maintz19981,LESTER1999129,hill2001medical,ZITOVA2003977,Holden2008review,Rueckert2011,sotiras2013deformable}. In the literature, many types of similarity metrics have been proposed for image registration, such as mean squared distance (MSD)~\cite{Woods1998auto,Hellier2002}, sum-of-squares distance (SSD)~\cite{THIRION1998243,Vercauteren08diffeomorphicdemons,Andersson2008Fnirt}, normalized cross correlation (CC)~\cite{ARDEKANI200567,Collins1997animal,shen2002hammer,cao2018region}, and normalized mutual information (MI)~\cite{Pluim03mutual,Maes1997multi,WELLS199635}. Besides, there are various regularization terms developed to penalize undesired deformations, such as topology preservation~\cite{Christensen2001cons,Rueckert2006diff,Musse2001topo,Musse2001topo,Sdika2008fast}, volume preservation~\cite{mansi2011ilogdemons,Tanner2000,GREENE2009809,Rohlfing2003,Bistoquet200869, Dauguet2009local}, and rigidity constraints~\cite{Dirk2004nonrigid,Maintz19981,Modersitzki2008}. In general, the existing registration algorithms can be roughly divided into two categories, including 1) non-learning-based methods, and 2) learning-based methods. We now introduce relevant studies in these two categories.

\subsection{Non-learning-based Registration Methods}
Non-learning-based registration algorithms typically optimize a transformation iteratively based on an energy function in the form of Eq.~\ref{eq_optimize}. Based on how to compute the similarity between the warped source image and the target image, there are two types of registration approaches, including 1) volume-based methods where the voxel intensities in the whole volume are used to drive the registration process, and 2) landmark-based methods where features extracted at anatomical landmarks are employed to guide the matching of the local correspondence during registration.

The most popular non-learning-based methods for deformable registration include automatic image registration (AIR)~\cite{Woods1998auto}, automatic registration toolbox (ART)~\cite{ARDEKANI200567}, HAMMER~\cite{shen2002hammer}, Demons~\cite{THIRION1998243}, diffeomorphic Demons ~\cite{Vercauteren08diffeomorphicdemons,lorenzi2013lcc}, statistical parametric mapping (SPM)~\cite{Hellier2002}, deformable registration via attribute matching and mutual-saliency weighting (DRAMMS)~\cite{Ou2009DRAMMSDR}, DROP~\cite{GLOCKER2008731}, CC/MI/SSD-FFD~\cite{rueckert1999nonrigid}, FNIRT~\cite{Andersson2008Fnirt}, and symmetric normalization (SyN)~\cite{avants2008symmetric}. Most of these methods require iterative optimization algorithms for parameter tuning~\cite{sotiras2013deformable}. Also, the registration performance of these methods would degrade when the source and target images have large variations in anatomical appearance. Therefore, robust and tuning-free deformable registration methods are highly desired for dealing with different data and registration tasks.

\subsection{Learning-based Registration Methods}
Many learning-based methods have been developed for deformable image registration~\cite{xue2006855}, such as those based on random forest~\cite{Wei2017learn}, support vector regression~\cite{Kim2012general}, sparse representation~\cite{Wang201561}, and deep neural networks~\cite{de2017end,balakrishnan2018unsupervised,cao2018deformable}. In these methods, deformable registration is often formulated as a learning problem to estimate the registration parameters. Compared with non-learning methods, learning-based approaches can predict the transformation efficiently for unseen testing images based on pre-trained models. According to whether supervision information is needed, existing learning-based methods for image registration can be further categorized into two types: 1) supervised learning methods, and 2) unsupervised learning methods.

\subsubsection{Supervised Methods}
In supervised methods for deformable image registration, task-specific supervision information (\eg, ground-truth registration) is usually required for model training. For instance, random forest has been applied for infant brain MRI registration and multi-modal image registration~\cite{Wei2017learn,Guti2016,Cao201718}, based on hand-engineered imaging features and pre-defined ground-truth registration. However, the registration performance of such traditional learning-based methods could be sub-optimal, since the process of feature extraction is independent of the model training.

Several deep-learning-based methods have been recently developed, by incorporating feature extraction and registration model learning into a unified framework. Cao~\etal~\cite{cao2018deformable} proposed a convolutional neural network (CNN) based regression model to directly learn the mapping from the input image pair (\ie, target and source images) to their corresponding deformation fields. Yang~\etal~\cite{YANG2017378} developed a fully convolutional network (FCN) to predict 3D deformable registration, followed by a correction network to further refine the predicted transformation. Roh{\'e}~\cite{Rohe2017} proposed an FCN model to learn the stationary velocity field, which consists of a contracting path to extract task-relevant features and a symmetric expanding path to output the transformation parameters. Sokooti~\etal~\cite{Sokooti2017nonrigid} proposed to predict displacement vectors by CNN models. Krebs~\etal~\cite{Krebs2017} adopted a deep reinforcement learning framework to estimate deformation fields. These methods usually require {\emph{task-specific ground-truth registration}} for model learning. Since ground-truth information is difficult to collect, supervised methods usually have limited utility in practice. Also, the performance of supervised methods is largely determined by the quality of predefined ground-truth registration.

\subsubsection{Unsupervised Methods}
By maintaining the unsupervised nature of deformable registration, unsupervised deep-learning methods have also been applied for medical image registration~\cite{wu2013unsupervised,miao2016cnn,shan2017unsupervised,de2017end,balakrishnan2018unsupervised}. That is, these methods do not rely on any pre-defined supervision information for network training. For example, Miao~\etal~\cite{miao2016cnn} developed a CNN model to learn transformation parameters for 2D/3D images, where imaging features for parameter regression should be predefined, which means this network cannot be trained in an end-to-end manner. Wu~\etal~\cite{wu2013unsupervised} proposed an unsupervised deep-learning algorithm for image registration based on image patches. Although this method can automatically extract features from images, it requires additional post-processing that cannot be handled inside CNNs.
Shan~\etal~\cite{shan2017unsupervised} developed an unsupervised end-to-end deep-learning model for 2D tissue registration, by directly predicting deformation field via a CNN.
In~\cite{de2017end}, an end-to-end unsupervised deep-learning model, consisting of a CNN-based regressor, a spatial transformer, and a re-sampler, was developed for deformable registration. Then, Guha~\etal~\cite{balakrishnan2018unsupervised} proposed an unsupervised CNN model, in which a spatial transformer network (STN) is also used to reconstruct one image from another while imposing smoothness constraints on the registration field. This method has achieved superior accuracy for 3D image registration in comparison to previous methods.

It is worth noting that most of the existing deep-learning methods ignore the inherent inverse-consistent property of transformations between a pair of images~\cite{shan2017unsupervised,de2017end,balakrishnan2018unsupervised}. That is, by independently estimating the transformation from an image $\A$ to an image $\B$ and that from $\B$ to $\A$, these methods are unable to ensure that these transforms are inverse mappings of one another. Note that several studies tackle this shortcoming by jointly estimating the transformations from both $\A$ to $\B$ and $\B$ to $\A$, under the consistency constraint that these transformations are inverses of one another~\cite{Christensen2001cons,leow2005inverse,he2003large}. However, these methods are not learning-based methods and require human-engineered feature representations (\eg, image intensity with Fourier series parameterization) of input images, which may not extracted in a task-oriented manner. Motivated by these studies, we develop an unsupervised deep network with an inverse-consistent constraint to encourage the inverse-consistent property of transformations between a pair of input images. Besides, considering using the smoothness constraint alone (as previous studies did) is not sufficient to guarantee that there is no folding in the estimated transformations, we further develop an anti-folding constraint to avoid foldings in the learned transformations.

\begin{figure*}[t]
\setlength{\belowdisplayskip}{0pt}
\setlength{\abovedisplayskip}{0pt}
\setlength{\abovecaptionskip}{-2pt}
\centering
\includegraphics[width=1\textwidth]{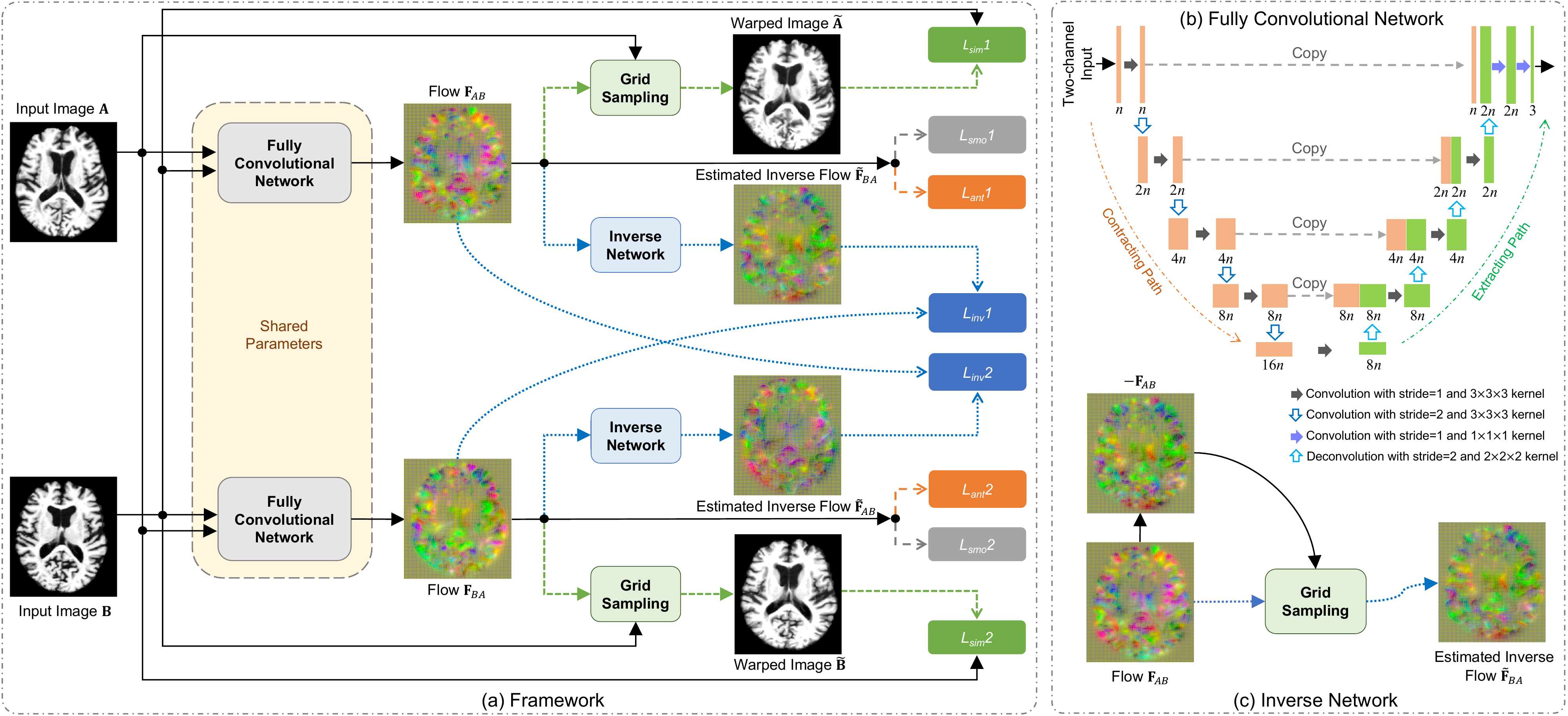}
\caption{Pipeline of the proposed Inverse-Consistent deep neural Network (ICNet) for unsupervised deformable image registration, which takes a pair of images as input. (a) Framework of ICNet, (b) architecture of fully convolutional network (FCN), and (c) illustration of the inverse network. The term $n$ in (b) denotes the number of starting convolutional channels in FCN.}
\label{fig_framework}
\end{figure*}

\section{Proposed Method}
\label{S3_method}
In this section, we first introduce the notations used in this paper. Then, we describe the proposed inverse-consistent deep neural network, as well as the objective function (with both the proposed inverse-consistent and anti-folding constraints) for network training. We finally introduce the implementation details for the proposed method.

\subsection{Notations}
In image registration, a pair of images are usually referred to as the source image and the target image. Since we do not rely on particular target images in our method, in this paper, we denote one input image as $\A$ and the other as $\B$. These two images are defined in the image domain $\Omega$. The transformation is a mapping function of the image domain $\Omega$ to itself, which spatially deforms any point locations to other locations. Also, we assume that the image $\A$ is deformed to match the image $\B$ according to a dense \emph{flow} (\ie, discrete displacement field) $\F_{AB}$ defined in the $\A$ space, while the image $\B$ is deformed to match the image $\A$ via another dense flow ${\F}_{BA}$ defined in the $\B$ space. Note that each element in a flow is a $3$-dimensional vector (corresponding to three axes, \ie, $x,y,z$), indicating the displacement of a particular voxel from its original location to a new location. In addition, the deformed/warped images of $\A$ and $\B$ are denoted as $\widetilde{\A}$ and $\widetilde{\B}$, respectively.
\if false
 are expressed as
\begin{equation}
\small
\begin{aligned}
\widetilde{\A}~ = &~ \Phi_\theta{(\A, \F_{AB})} \\
\widetilde{\B}~ = &~\Phi_\theta{(\B, \F_{BA})}
\end{aligned}
\end{equation}
where $\Phi_\theta{(\A, \F_{AB})}$ denotes the transformation operation from the image $\A$ to its warped image $\widetilde{\A}$ via the flow $\F_{AB}$, and $\Phi_\theta{(\B, \F_{BA})}$ is the transformation operation from the image $\B$ to the warped image $\widetilde{\B}$ using the flow $\F_{BA}$. Here, $\theta$ represents the to-be-estimated transformation parameters.
\fi

\subsection{Inverse-consistent Unsupervised Neural Network}
Figure~\ref{fig_framework} illustrates the proposed unsupervised deep network for deformable image registration. As shown in Fig.~\ref{fig_framework}~(a), we employ a fully convolutional network (FCN) to model two dense, non-linear transformations (\ie, $\F_{AB}$ and $\F_{BA}$) from a pair of input images (\ie, $\A$ and $\B$) to their warped images (\ie, $\widetilde{\A}$ and $\widetilde{\B}$). There are two FCN modules in our proposed network. The first one is used to align the image $\A$ to $\B$ (as the target image) using the flow $\F_{AB}$, generating the warped image $\widetilde{\A}$. In contrast, the second FCN is designed to model the registration mapping from the image $\B$ to $\A$ (as the target image) via the flow $\F_{BA}$, yielding the warped image $\widetilde{\B}$. It is worth noting that these two FNCs share network structure and parameters.

As shown in Fig.~\ref{fig_framework}~(b), the FCN we used here follows a U-Net architecture~\cite{ronneberger2015u} to capture and combine both global and local structural information of input images. Specifically, the input data contain two channels (with each channel corresponds to a particular input image), and $n$ is the number of starting filter channels of the FCN. The FCN contains a contracting path for image down-sampling and an expanding path for image up-sampling. Every step in the contracting path contains a {\small$3\times3\times3$} convolution with a stride of $1$, and a {\small$3\times3\times3$} convolution with a stride of $2$ for down-sampling. Besides, each step in the expanding path consists of a {\small$2\times2\times2$} deconvolution with a stride of $2$ for up-sampling, followed by a concatenation process to combine up-sampled feature maps with the corresponding feature maps from the contracting path, and then a {\small$3\times3\times3$} convolution with a stride of $1$. In this network, each convolution is followed by a rectified linear unit (ReLU) activation, while the output of the last layer (having $3$ filter channels that are corresponding to the $x$, $y$, and $z$ axis) is constrained into $[-\tau,\tau]$. That is, we first use a $tanh$ function to normalize the output of the last layer to $[-1,1]$, followed by multiplying by a constant $\tau$ (\ie, the maximum displacement magnitude). Since both input images are treated equally in the proposed ICNet framework, the two FCNs in Fig.~\ref{fig_framework}~(b) share the same parameters.

Besides, we can see from Fig.~\ref{fig_framework}~(a) that a grid sampling module is utilized to generate the warped image (\eg, $\widetilde{\A}$), based on the input image (\eg, $\A$) and the learned flow (\eg, $\F_{AB}$). Specifically, such grid sampling is implemented via the fully-differentiable spatial transformer network (STN)~\cite{jaderberg2015spatial}, containing a regular spatial grid generator and a sampler. The flow (displacement field) predicted by our image registration network is used to transform the regular spatial grid into a sampling grid. Then, the sampler uses the sampling grid to warp the input image. Bilinear interpolation is used during the sampling process, making STN fully differentiable for back propagation.

Furthermore, an inverse network is developed to generate an estimated inverse flow (\eg, $\widetilde{\F}_{BA}$) of each transformation learned by the FCN (\eg, ${\F}_{AB}$), based on which an inverse-consistent loss (\ie, $Loss_{inv}1$ and $Loss_{inv}2$ ) is further adopted to encourage the inverse-consistent property of two transformations. As shown in Fig.~\ref{fig_framework}~(c), we utilize the grid sampling strategy to generate the estimated inverse flow $\widetilde{\F}_{BA}$, based on both ${\F}_{AB}$ and ${-\F}_{AB}$.

\subsection{Proposed Objective Function}

\subsubsection{Inverse-consistent Constraint}
Existing deep-learning methods typically ignore the inverse-consistent property of transformations between a pair of images~\cite{shan2017unsupervised,de2017end,balakrishnan2018unsupervised}. Motivated by previous non-learning-based inverse-consistent methods~\cite{Christensen2001cons,leow2005inverse,he2003large}, we propose to simultaneously estimate the transformation from $\A$ to $\B$ (\ie, $\F_{AB}$) and the transformation from $\B$ to $\A$ (\ie, $\F_{AB}$), and enforce the consistency constraint that these bidirectional transformations are inverse mappings of one another.

Specifically, we propose an \emph{inverse-consistent regularization} term to penalize the difference between two transformations from the respective inverse mappings. As shown in Fig.~\ref{fig_framework}~(c), we rely on an inverse network to generate the inverse mapping (\eg, $\widetilde{\F}_{BA}$) of each transformation (\eg, ${\F_{AB}}$). Specifically, for the flow $\F_{AB}$, we first obtain its negative flow (\ie, $-{\F_{AB}}$) in the $\A$ space. We then feed both $\F_{AB}$ and $-\F_{AB}$ to the grid sampling module (via a STN) to obtain the estimated inverse flow (\ie, $\widetilde{\F}_{BA}$ in the $\B$ space) of $\F_{AB}$. Similarly, we feed both $\F_{BA}$ and its negative flow $-\F_{BA}$ to the grid sampling module, and hence can obtain the estimated inverse flow (\ie, $\widetilde{\F}_{AB}$) of $\F_{BA}$. Mathematically, the proposed inverse-consistent constraint can be defined as follows
\begin{equation}
\small
\begin{aligned}
\mathcal{L}_{inv} = {\parallel \F_{AB}-\widetilde{\F}_{AB} \parallel}_F^2 + {\parallel \F_{BA}-\widetilde{\F}_{BA} \parallel}_F^2
\end{aligned}
\label{eq_inverseLoss}
\end{equation}
with
\begin{equation}
\small
\begin{aligned}
\widetilde{\F}_{AB} = & \mathcal{G} \left(\F_{BA}, -\F_{BA}  \right) \\
\widetilde{\F}_{BA} = & \mathcal{G} \left(\F_{AB}, -\F_{AB}  \right)
\end{aligned}
\label{eq_inverseFlow}
\end{equation}
where $\mathcal{G}$ is the mapping generated by the grid sampling module (via a STN), and ${\parallel \cdot \parallel}_F$ represents the Frobenius norm of a matrix. The two terms in Eq.~\ref{eq_inverseLoss} correspond to the notations $L_{inv}1$ and $L_{inv}2$ in Fig.~\ref{fig_framework}~(a). By minimizing Eq.~\ref{eq_inverseLoss}, we concurrently encourage both the difference between the flows $\F_{AB}$ and $\widetilde{\F}_{AB}$ (\ie, the inverse of $\F_{BA}$) and that between $\F_{BA}$ and $\widetilde{\F}_{BA}$ (\ie, the inverse of $\F_{AB}$) to be small. In this way, the inverse-consistent property of the to-be-estimated transformations can be explicitly modeled in the proposed network.

\subsubsection{Anti-folding Constraint}
As mentioned in Section~\ref{Sec:introduction} (\eg, Fig.~\ref{fig_smootEffect}), if we excessively encourage local smoothness of the to-be-estimated flow by using a large weight for the smoothness constraint, the registration results will be inaccurate. Otherwise, there will be possible foldings in the flow, thus yielding unreasonable registration. To deal with this issue, besides using the conventional smoothness constraint, we also develop an anti-folding constraint as
\if false
\begin{equation}
\small
\begin{aligned}
\mathcal{L}_{ant} = \Sigma_{p \in \Omega} {
\,\delta{ \left( \nabla{\F_{AB}\left(p\right)}+1  \right)}  {\parallel {\nabla{\F_{AB}}\left(p \right)}\parallel}_2^2   }  \\
+ {\delta{ \left( \nabla{\F_{BA}\left(p\right)}+1  \right)} {\parallel {\nabla{\F_{BA} \left(p \right)}} \parallel}_2^2 }
\end{aligned}
\label{eq_lossAnti}
\end{equation}
\fi
\begin{equation}
\small
\begin{aligned}
\mathcal{L}_{ant} = \Sigma_{p \in \Omega}  \Sigma_{i \in \lbrace{x,y,z}\rbrace} {
\,\delta{ \left( \nabla{\F^i_{AB}\left(p\right)}+1  \right)}  {\vert{\nabla{\F^i_{AB}}\left(p \right)}\vert}^2   }  \\
+ \,{\delta{ \left( \nabla{\F^i_{BA}\left(p\right)}+1  \right)} {\vert{\nabla{\F^i_{BA} \left(p \right)}} \vert}^2 }
\end{aligned}
\label{eq_lossAnti}
\end{equation}
where ${\nabla{\F^i_{AB} \left(p \right)}}$ is the gradient of the flow ${\F_{AB}}$ along the $i$-th ($i \in \lbrace{x,y,z}\rbrace$) axis at the location of the voxel $p$. Besides, the term $\delta{\left(Q\right)}$ is an index function used to penalize the gradient of the flow at the locations with foldings. That is, if $Q\le0$, $\delta{\left( Q \right)} = |Q|$; and $\delta{\left(Q\right)}=0$, otherwise.

The purpose of Eq.~\ref{eq_lossAnti} can be explained as follows. If there is a folding at the location of $p$ along the $i$-th axis (\ie, $\nabla{\F^i_{AB}\left(p\right)}+1\le0$), we enforce the penalty $|\nabla{\F^i_{AB}\left(p\right)}+1|$ on the gradient at this location, requiring this gradient to be small. In contrast, if $\nabla{\F^i_{AB}\left(p\right)}+1>0$ (\ie, no folding at the location of $p$ along the $i$-th axis), we do not penalize the gradient at this location. More detailed explanation can be found in the \emph{Appendix}.

\subsubsection{Smoothness Constraint}
In previous studies, the to-be-estimated deformation field is generally to be locally smoothed via a smoothness constraint on its spatial gradients~\cite{de2017end,balakrishnan2018unsupervised}. Here, we also use such smoothness constraint in the objective function as
\begin{equation}
\small
\begin{aligned}
\mathcal{L}_{smo} = \Sigma_{p \in \Omega} {\left({\parallel \nabla {\F_{AB}\left(p\right)}  \parallel}_2^2+ {\parallel  \nabla {\F_{BA}\left(p\right)}   \parallel}_2^2 \right)}
\end{aligned}
\label{eq_smooth}
\end{equation}
where $\nabla {\F_{AB}\left(p\right)} $ is the gradient of the flow $\F_{AB}$ at the voxel $p$, while $\nabla {\F_{BA}\left(p\right)} $ denotes the gradient of the flow $\F_{BA}$ at the voxel $p$. The operation ${\parallel \cdot \parallel}_2$ represents the $l_2$ norm of a vector. Here, we approximate the spatial gradients using the differences between neighboring voxels.

\subsubsection{Objective Function}
In this work, we utilize the mean squared distance (MSD) as the similarity metric to compare the alignment between the warped image and its corresponding target image. Specifically, the MSD-based symmetric similarity is employed to measure the shape differences between the deformed image $\widetilde{\A}$ and image $\B$ and the differences between the deformed image $\widetilde{\B}$ and image $\A$, which is defined as
\begin{equation}
\small
\begin{aligned}
\mathcal{L}_{sim} =  {{\parallel \B - \widetilde{\A} \parallel}_F^2 + {\parallel \A - \widetilde{\B} \parallel}_F^2 }
\end{aligned}
\label{eq_lossAll}
\end{equation}
where $\widetilde{\A} = \mathcal{G}\left( \F_{AB}, \A \right)$ and $\widetilde{\B} = \mathcal{G}\left( \F_{BA}, \B \right)$, and $\mathcal{G}$ is the mapping function learned in the grid sampling module.

Accordingly, the loss function of our proposed ICNet for deformable registration is formulated as follows
\begin{equation}
\begin{aligned}
 \mathcal{L}(\A,\B)=~& \mathcal{L}_{sim} + \alpha \mathcal{L}_{smo} \\
+ ~&\beta \mathcal{L}_{inv} + \gamma \mathcal{L}_{ant} \\
\end{aligned}
\label{Eq_loss}
\end{equation}
where the parameters $\alpha$, $\beta$ and $\gamma$ are used to balance the contributions of the smoothness, inverse-consistent and anti-folding regularizers, respectively.

\subsection{Implementation}
\begin{figure*}[t]
	\setlength{\belowdisplayskip}{0pt}
	\setlength{\abovedisplayskip}{0pt}
	\setlength{\abovecaptionskip}{-2pt}
	\centering
	\includegraphics[width=1\textwidth]{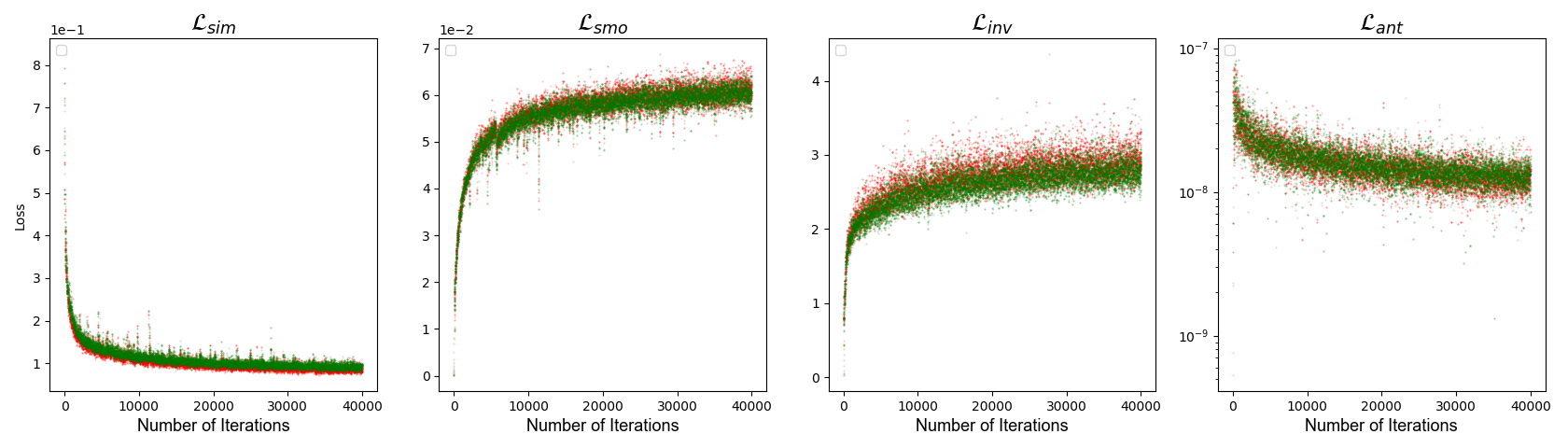}
	\caption{Training and validation loss of the proposed ICNet regarding (a) similarity term, (b) inverse-constraint regularizer, (c) anti-folding regularizer, and (d) smooth regularizer. Red lines denote the training loss, while green lines correspond to the validation loss.}
	\label{fig_trainValidationLoss}
\end{figure*}

We implement the proposed deep network with Pytorch~\cite{ketkar2017introduction}. The objective function in Eq.~\ref{Eq_loss} is optimized by the Adam algorithm~\cite{kingma2014adam} combined with a back-propagation algorithm for computing gradients as well as updating network parameters. The learning rate for Adam is empirically set to $5\times10^{-4}$. In Fig.~\ref{fig_trainValidationLoss}, we report the change curves of both training (red) and validation (green) losses on the public Alzheimer's Disease Neuroimaging Initiative (ADNI-1) database~\cite{jack2008alzheimer}, where $10$\% subjects are randomly selected as the validation data, and the remaining subjects are used as the training data. This figure indicates that the proposed ICNet method generalizes well with almost no over-fitting issue, and the proposed objective functions converges quickly within $40,000$ iterations. For readers' convenience, our code and trained model will be made publicly available online.

\section{Experiments}
\label{S4_experiments}
In this section, we first introduce the studied materials, competing methods, and experimental settings. Then, we present results of brain tissue segmentation and anatomical landmark detection based on the warped MR images achieved by different registration methods. We finally analyze the computational costs of different methods.

\subsection{Materials and Image Pre-processing}
We perform experiments on $860$ subjects from two subsets of the public ADNI database\footnote{\url{http://adni.loni.usc.edu}}~\cite{jack2008alzheimer}, \ie, 1) ADNI-1, and 2) ADNI-2. To be specific, there are $805$ subjects with the baseline structural brain MRI scans in ADNI-1, while the remaining $55$ subjects with the baseline structural MRI data are randomly selected from ADNI-2. Since several subjects participated in both ADNI-1 and ADNI-2, we simply remove these subjects from ADNI-2, to ensure that ADNI-1 and ADNI-2 are independent datasets in the experiments. Notably, the studied subjects from ADNI-1 have $1.5\,$T T1-weighted MRI scans, while those in ADNI-2 have $3.0\,$T T1-weighted MRI data.

For all structural brain MR images, we perform both \emph{spatial normalization} and \emph{intensity normalization} for image normalization. For spatial normalization, we first perform skull stripping and cerebellar removal for all brain MRIs, and then linearly align them to a common Colin27~\cite{holmes1998enhancement} template. We further resample all linearly aligned images to have the same spatial resolution (\ie, $1\,mm\times1\,mm\times1\,mm$), followed by cropping them to have the same image size (\ie, $144\times192\times160$). For intensity normalization, we first match the intensity histogram of each brain MRI to that of the Colin27 template by using a histogram matching algorithm. Then, we also perform the z-score normalization to make the mean intensity of each image is zero and the standard deviation is one.

\begin{figure}[htbp]
	\setlength{\belowdisplayskip}{0pt}
	\setlength{\abovedisplayskip}{0pt}
	\setlength{\abovecaptionskip}{-2pt}
	\centering
	\includegraphics[width=0.49\textwidth]{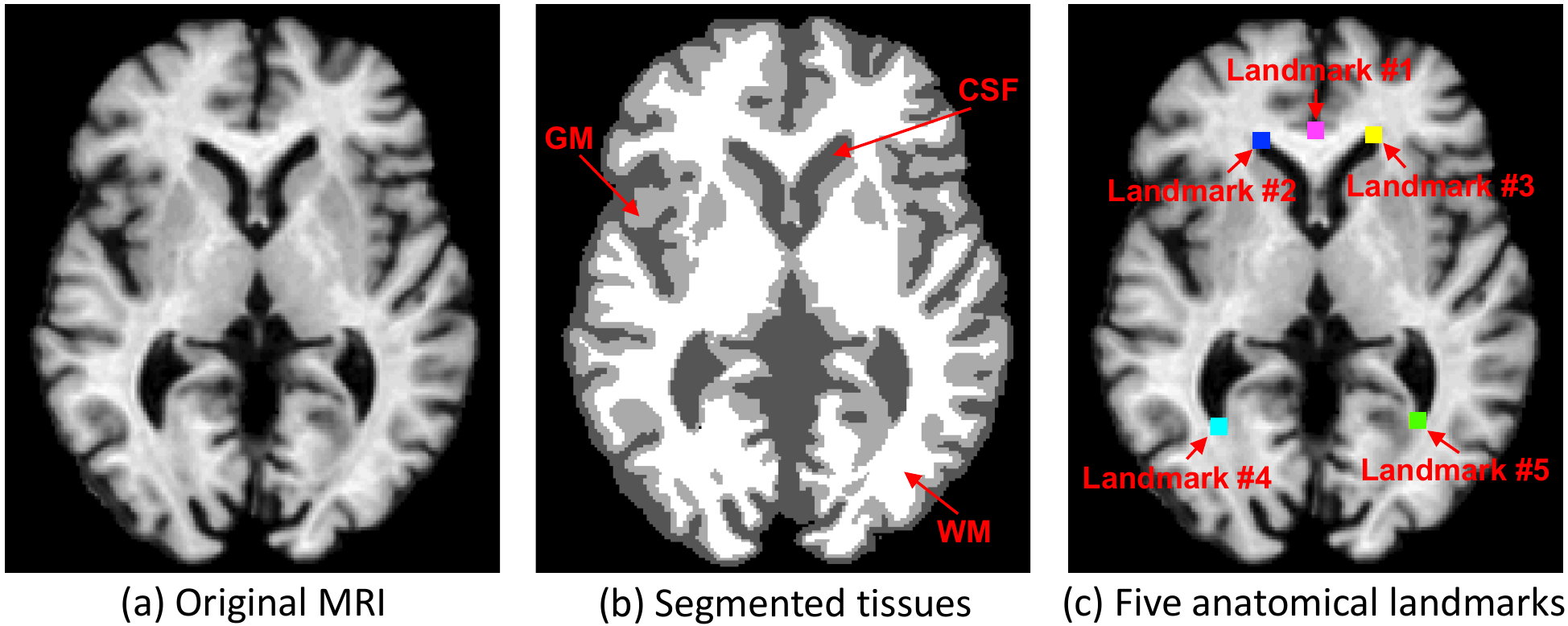}
	\caption{Illustration of ground-truth tissue segmentations and anatomical landmarks: (a) the original MRI, (b) three types of segmented tissues (\ie, CSF, GM, and WM), and (c) five anatomical landmarks denoted by blocks with different colors.}
	\label{fig_groundtruth}
\end{figure}

In the experiments, we perform two tasks to evaluate the registration performance of different methods, including 1) \emph{brain tissue segmentation} and 2) \emph{anatomical landmark detection}. In the task of brain tissue segmentation, the ground-truth segmentation is generated by using first FAST in FSL~\cite{zhang2001segmentation} to obtain the tissue segmentation map, followed by further manual correction. As illustrated in Fig.~\ref{fig_groundtruth}~(b), three tissues are segmented from each brain MR image, including cerebrospinal fluid (CSF), gray matter (GM) and white matter (WM). In the task of anatomical landmark detection, the ground-truth landmarks are manually annotated by an experienced radiologist. As shown in Fig.~\ref{fig_groundtruth}~(c), five anatomical landmarks are annotated in each brain MR image, which are mainly located in ventricles.

\subsection{Competing Methods}
We compare the proposed {\bf{ICNet}} method with three state-of-the-art methods for deformable image registration, including 1) {\bf{Demons}} with the symmetric local correlation coefficient used as the similarity metric~\cite{lorenzi2013lcc}, 2) symmetric normalization (denoted as {\bf{SyN}})~\cite{avants2008symmetric}, and 3) a unsupervised deep-learning (denoted as {\bf{DL}}) method with a minimum mean squared error (MMSE) loss and a smoothness constraint~\cite{balakrishnan2018unsupervised}. Among them, Demons and SyN are non-learning-based methods, while DL is an unsupervised learning method. For the fair comparison, the network architecture of the DL method is similar to our ICNet, while its objective function is different from ours. Specifically, only the first two terms in Eq.~\ref{Eq_loss} is included in the objective function of DL, and hence, DL can be regarded as a degenerated variant of ICNet.

{
\makegapedcells
\setcellgapes{1.1pt}
	
\begin{table*}[!tp]
    \setlength{\belowcaptionskip}{-1pt}
    \setlength{\abovecaptionskip}{-1pt}
    \setlength\abovedisplayskip{-2pt}
    \setlength\belowdisplayskip{-4pt}
    \renewcommand\arraystretch{1.2}
    \centering
    \caption{Segmentation results for three brain tissues (\ie, CSF, GM and WM), achieved by six different methods on the ADNI dataset.}
    \scriptsize
\begin{tabular*}{0.98\textwidth}{@{\extracolsep{\fill}} cl cccc cc }
\toprule
{~~Tissue}  & Metric  &   Demons  &    SyN     &     DL &  ICNet-1  &ICNet-2  &ICNet \\
\midrule
\multicolumn{1}{l}{ \multirow{5}{*}{CSF} }
& DSC (\%) &  $77.16\pm1.82$  & $78.43\pm1.38$ & $79.76\pm1.22$  & $80.75\pm1.31$  & $82.08\pm1.14$ & $\bm{83.58\pm1.17}$ \\
& SEN (\%) &  $71.99\pm2.15$  & $76.57\pm1.96$ & $77.38\pm1.94$  & $79.00\pm1.84$  & $80.16\pm1.75$ & $\bm{82.55\pm1.80}$ \\
& PPV (\%) &  $83.16\pm2.07$  & $80.41\pm1.47$ & $82.34\pm1.38$. & $82.61\pm1.44$  & $84.13\pm1.36$ & $\bm{84.66\pm1.31}$ \\
& ASD &  $0.69\pm0.05$    & $0.73\pm0.03$   & $0.68\pm0.03$    & $ 0.65\pm0.03$   & $0.60\pm0.03$   & $\bm{0.56\pm0.03}$ \\
& HD   & $11.63\pm0.84$ & $11.74\pm0.78$ & $11.60\pm0.80$  & $11.41\pm0.74$   & $11.29\pm0.71$ & $\bm{11.10\pm0.78}$ \\
\midrule
\multicolumn{1}{l}{ \multirow{5}{*}{GM} }
& DSC (\%) & $76.15\pm1.33 $ & $74.16\pm1.10$ & $75.04\pm1.06$ & $77.93\pm1.05$ & $78.30\pm1.04$ & $\bm{80.59\pm0.98}$ \\
& SEN (\%) & $74.71\pm1.63 $ & $74.23\pm1.20$ & $74.41\pm1.22$ & $77.96\pm1.24$ & $77.92\pm1.25$ & $\bm{82.28\pm1.15}$  \\
& PPV (\%) & $77.65\pm1.38 $ & $74.11\pm1.44$ & $75.70\pm1.41$ & $77.92\pm1.38$ & $78.70\pm1.37$ & $\bm{78.99\pm1.38}$\\
& ASD & $0.58\pm0.03   $ & $0.65\pm0.02$   & $0.63\pm0.02$   & $0.58\pm0.02$   & $0.56\pm0.02$   & $\bm{0.54\pm0.02}$ \\
& HD   & $9.85\pm1.25  $  & $9.13\pm0.92$   & $9.13\pm1.01$   & $9.20\pm1.04$  & $\bm{8.95\pm0.95}$    & $9.07\pm1.05$\\
\midrule
\multicolumn{1}{l}{ \multirow{5}{*}{WM} }
& DSC (\%) & $84.02\pm1.75$ & $84.67\pm1.10$ & $86.03\pm1.01$ & $86.92\pm1.11$  & $87.82\pm1.00$ & $\bm{88.13\pm0.94}$\\
& SEN (\%) & $80.99\pm2.04$ & $82.79\pm1.75$ & $85.35\pm1.63$ & $85.03\pm1.73$  & $86.17\pm1.71$ & $\bm{86.56\pm1.61}$\\
& PPV (\%) & $87.31\pm2.05$ & $86.66\pm1.25$ & $86.74\pm1.24$ & $ 88.92\pm1.32$ & $89.13\pm1.24$ & $\bm{90.22\pm1.19}$\\
& ASD & $0.80\pm0.06 $  & $0.88\pm0.04$   & $0.81\pm0.04$   & $0.76\pm0.04$    & $\bm{0.71\pm0.04}$   & $\bm{0.71\pm0.04}$\\
& HD   & $13.73\pm1.38$ & $13.40\pm1.13$ & $12.77\pm1.22$ & $13.18\pm1.28$  & $13.16\pm1.24$ & $\bm{12.71\pm1.11}$ \\

\bottomrule
\end{tabular*}
\label{tab_resultsADNI}
\end{table*}
}

\begin{figure*}[htbp]
	\setlength{\belowdisplayskip}{0pt}
	\setlength{\abovedisplayskip}{0pt}
	\setlength{\abovecaptionskip}{-2pt}
	\centering
	\includegraphics[width=1\textwidth]{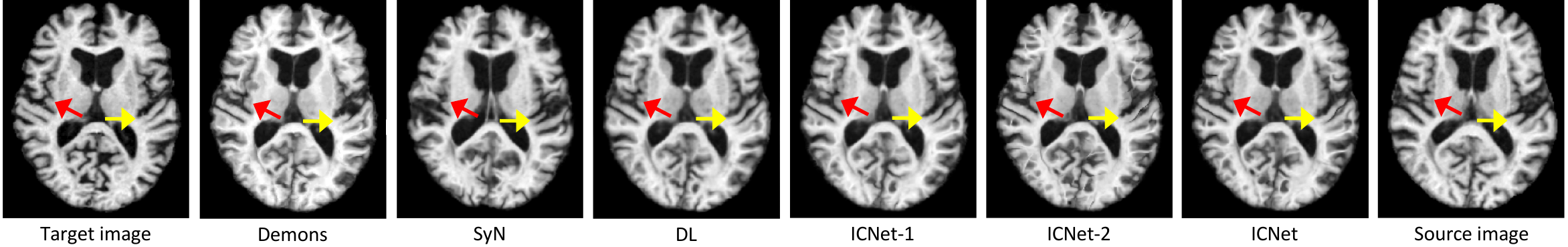}
	\caption{Registration results achieved by six different methods for deforming a source image (right) to a target image (left). Red and yellow arrows indicate the right and left planum temporale, respectively.}
	\label{fig_visualResults}
\end{figure*}

To evaluate the effectiveness of the proposed two constraints (\ie, the inverse-consistent constraint and the anti-folding constraint), we further compare ICNet with its two variants. The first variant is denoted as {\bf{ICNet-1}}, in which the proposed inverse-consistent constraint is removed. Similarly, the second variant is denoted as {\bf{ICNet-2}}, in which the proposed anti-folding constraint is not used.

\subsection{Experimental Settings}
Since the six methods for comparison are all unsupervised, we do not need to generate any ground-truth registration for them. For the non-learning-based methods (\ie, Demons and SyN), we utilize their recommended parameter settings in the experiments. For the learning-based methods (\ie, DL, ICNet-1, ICNet-2, and ICNet), we treat the ADNI-1 and ADNI-2 as the training set and testing set, respectively. We randomly select $10\%$ subjects from the ADNI-1 as the validation data, while the remaining subjects are used as the training data.

In the {\emph{training}} stage, we randomly select a pair of MR images from the ADNI-1 as the input for each learning-based method. In the {\emph{testing}} stage for all competing methods, to perform deformable registration, we first select $5$ MR images from the ADNI-2 as atlas images, while the remaining $50$ MR images are used as testing images. By using different deformable registration algorithms, we first warp each of the $5$ atlases to a particular testing image, and thus generating $5$ warped atlas images based on this testing image. Then, we employ a multi-atlas based segmentation algorithm with a majority voting strategy~\cite{aljabar2009multi} to perform brain tissue segmentation in each testing image. Similarly, for landmark detection, each landmark in the atlases is first mapped onto a particular testing image via the corresponding deformable transformation~\cite{zhang2017detecting}. Thus, given a testing image, we obtain $5$ warped landmark positions for each landmark, followed by averaging these positions to generate the final location of this landmark. It is worth noting that, to evaluate the performance of the proposed ICNet for deformable image registration, we only utilize atlas-based methods for tissue segmentation and landmark detection, while other supervised learning methods are beyond the scope of this paper.

In ICNet, the parameter $\gamma$ is empirically set to $10^5$ to avoid folding in the flow as much as possible, and the other two parameters (\ie, $\alpha$ and $\beta$) are determined via grid search within the range of $[10^{-5}, 10^{-4},\cdots,10^{5}]$ on the validation set. Similarly, parameters in the three competing methods (\ie, $\alpha$ in DL, $\alpha$ in ICNet-1, and $\alpha, \beta$ in ICNet-2) are also selected from the same range through cross-validation. Besides, we set $\beta=0, \gamma=10^5$ in ICNet-1, and $\gamma=0$ in ICNet-2. The number of starting filter channels $n$ (see Fig.~\ref{fig_framework}~(b)) for ICNet and its two variants (\ie, ICNet-1 and ICNet-2) is empirically set to $8$. In each of the four deep-learning methods (\ie, DL, ICNet-1, ICNet-2, and ICNet), the output of the last layer (with $3$ filter channels corresponding to the $x$, $y$, and $z$ axis) is constrained into the range $[-\tau,\tau]$. Following~\cite{cao2017deformable}, we empirically set $\tau=7$ in this work, considering the displacement magnitude is usually less than $7$ in the deformable registration of brain MRI scans.

{
\makegapedcells
\setcellgapes{1.2pt}	
\begin{table*}[!tp]
    \setlength{\belowcaptionskip}{-1pt}
    \setlength{\abovecaptionskip}{-1pt}
    \setlength\abovedisplayskip{-1pt}
    \setlength\belowdisplayskip{-1pt}
    \renewcommand\arraystretch{1.2}
    \centering
    \caption{Landmark detection error ($mm$) achieved by six different methods on the ADNI dataset.}
 \scriptsize
\begin{tabular*}{0.96\textwidth}{@{\extracolsep{\fill}} l cc cc cc }
\toprule
~~Index &   Demons  &    SyN     &     DL &  ICNet-1  &ICNet-2  &ICNet \\
\midrule
~~Landmark \#$1$ &  $2.40\pm0.96$ &  $2.44\pm0.96$ &  $2.18\pm0.87$ &  $2.19\pm0.83$ &  $2.12\pm0.91$ &  $\bm{2.10\pm0.91}$ \\
~~Landmark \#$2$ &  $3.07\pm1.35$ &  $2.77\pm1.26$ &  $2.65\pm1.14$ &  $\bm{2.53\pm1.07}$ &  $2.66\pm1.15$ &  $2.68\pm1.16$ \\
~~Landmark \#$3$ &  $3.61\pm1.84$ &  $3.56\pm1.78$ &  $3.02\pm1.69$ &  $3.03\pm1.80$ &  $2.75\pm1.63$ &  $\bm{2.60\pm1.57}$ \\
~~Landmark \#$4$ &  $3.41\pm1.30$ &  $3.23\pm1.43$ &  $2.78\pm1.29$ &  $2.82\pm1.36$ &  $\bm{2.72\pm1.18}$ &  $2.73\pm1.16$ \\
~~Landmark \#$5$ &  $3.32\pm1.21$ &  $3.41\pm1.34$ &  $3.06\pm1.19$ &  $3.06\pm1.25$ &  $2.95\pm1.15$ &  $\bm{2.94\pm1.14}$ \\
~~Average               &  $3.16\pm1.42$ &  $3.08\pm1.44$ &  $2.74\pm1.31$ &  $2.73\pm1.34$ &  $2.64\pm1.26$ &  $\bm{2.61\pm1.24}$ \\

\bottomrule
\end{tabular*}
\label{tab_resultsLandmark}
\end{table*}
}

In the experiments of brain tissue segmentation, five complementary metrics are used for quantitative evaluation of segmentation performance, including 1) dice similarity coefficient (DSC), 2) sensitivity (SEN), 3) positive predictive value (PPV), 4) average symmetric surface distance (ASD), and 5) Hausdorff distance (HD). In the experiments of anatomical landmark detection, for each landmark, we report the landmark detection error by computing the Euclidean distance between the estimated landmark location (achieved by a specific method) and its ground-truth location. For the evaluation metrics of ACC, SEN and PPV, higher values indicate better performance. For the remaining three metrics (\ie, ASD, HD, and detection error), lower values denote better performance.

\subsection{Results of Brain Tissue Segmentation}
In the first group of experiments, we perform the segmentation of three types of brain tissues (\ie, CSF, GM and WM), based on the warped atlas images generated by six different methods. The experimental results are shown in Table~\ref{tab_resultsADNI}.

From Table~\ref{tab_resultsADNI}, one could have the following observations. \emph{First}, in most cases, the proposed methods (\ie, ICNet-1, ICNet-2, and ICNet) achieve the overall best performance (regarding DSC, SEN, PPV, ASD, and HD) for segmenting all the three types of tissues. For instance, the DSC value achieved by ICNet for CSF segmentation is $83.58\%$, while the DSC produced by the conventional deep-learning method (\ie, DL) is only $79.76\%$. \emph{Second}, even though no supervision information is required, four unsupervised deep-learning-based methods (\ie, DL, ICNet-1, ICNet-2, and ICNet) usually outperform two non-learning-based methods (\ie, Demons and SyN). The underlying reason may be that deep-learning-based methods can extract task-oriented features via neural networks, while conventional methods simply employ hand-engineered features of brain MRIs. \emph{Besides}, we can see that our ICNet method usually outperform its two variants (\ie, ICNet-1 and ICNet-2). Note that ICNet-1 does not use the proposed inverse-consistent constraint, and ICNet-2 does not utilize the proposed anti-folding constraint. This implies that including both the inverse-consistent and anti-folding constraints to train our ICNet could boost the deformable image registration performance.

Given a testing image, we further visually compare the registration results for a source brain MR image achieved by different methods in Fig.~\ref{fig_visualResults}. From Fig.~\ref{fig_visualResults}, we can see that the proposed ICNet method brings impressive improvement for the registration results, compared with the competing methods. For instance, it is obvious that the regions of the left and right planum temporale are more accurately registered to the target image using ICNet, as indicated by the red arrow (left planum temporale) and the yellow arrow (right planum temporale) in Fig.~\ref{fig_visualResults}.

\subsection{Results of Anatomical Landmark Detection}
In the second group of experiments, we perform landmark detection based on the deformed atlas images generated by different registration methods, with the results reported in Table~\ref{tab_resultsLandmark}. From Table~\ref{tab_resultsLandmark}, we can see that the overall performance of our ICNet method is superior to the five competing methods. For instance, the average landmark detection error achieved by ICNet is $2.61$, which is lower than the result of Demons (\ie, $3.16$).

It is worth noting that our methods (\ie, ICNet-1, ICNet-2, and ICNet) are unsupervised and do not need to generate the ground-truth flow for each to-be-registered image during network training. This is a particularly useful property for the deformable registration algorithm, which not only maintains the unsupervised nature of deformable registration, but also avoids the challenge of collecting accurate ground-real registration.

\subsection{Computational Cost}
We now analyze the computational costs of the proposed ICNet method and those competing methods. For the four deep-learning-based methods (\ie, DL, ICNet-1, ICNet-2, and ICNet), the training process is performed off-line, while the non-leanding-based methods (\ie, Demons and SyN) do not need any training process. Hence, we only analyze the on-line computational cost for nonlinearly registering a new brain MRI in the application/testing stage. Table~\ref{tab_computeCost} reports the computational costs of different methods. Note that Demons and SyN are implemented using a CPU (i7-7700, 3.6GHz), while the remaining four methods are implemented using a GPU (GTX 1080ti).
We can observe from Table~\ref{tab_computeCost} that the computational costs of the four deep-learning-based methods require only $\sim 0.25$ second for deformable registration of one MRI, which is much faster than Demons ($\sim 1$ minutes) and SyN ($\sim 2$ hours). These results further demonstrate the potential utility of our method in practical applications.

{
\makegapedcells
\setcellgapes{1.3pt}	
\begin{table}[htbp]
    \setlength{\belowcaptionskip}{-1pt}
    \setlength{\abovecaptionskip}{-1pt}
    \setlength\abovedisplayskip{-2pt}
    \setlength\belowdisplayskip{-4pt}
    \renewcommand\arraystretch{1.2}
    \centering
    \caption{Computation time of different methods for deformable registration of each testing image in the application stage.}
 \scriptsize
\begin{tabular*}{0.48\textwidth}{@{\extracolsep{\fill}} l c cc ccc }
\toprule
~Method &   Demons  &    SyN     &     DL  &ICNet-1 &ICNet-2 &ICNet \\
\midrule
~Time &  $\sim1m$ &  $\sim2h$ &  $\sim0.25s$ &  $\sim0.25s$ &  $\sim0.25s$ &  $\sim0.25s$ \\
\bottomrule
\end{tabular*}
\label{tab_computeCost}
\end{table}
}

\section{Discussion}
\label{S5_discussion}
In this section, we first investigate the effect of two essential components (\ie, the inverse-consistent and anti-folding constraints) in the proposed ICNet method. We then analyze the influence of network architectures and a network refining strategy used in the application stage.

\begin{figure*}[t]
\setlength{\belowdisplayskip}{0pt}
\setlength{\abovedisplayskip}{0pt}
\setlength{\abovecaptionskip}{-2pt}
\centering
\includegraphics[width=1\textwidth]{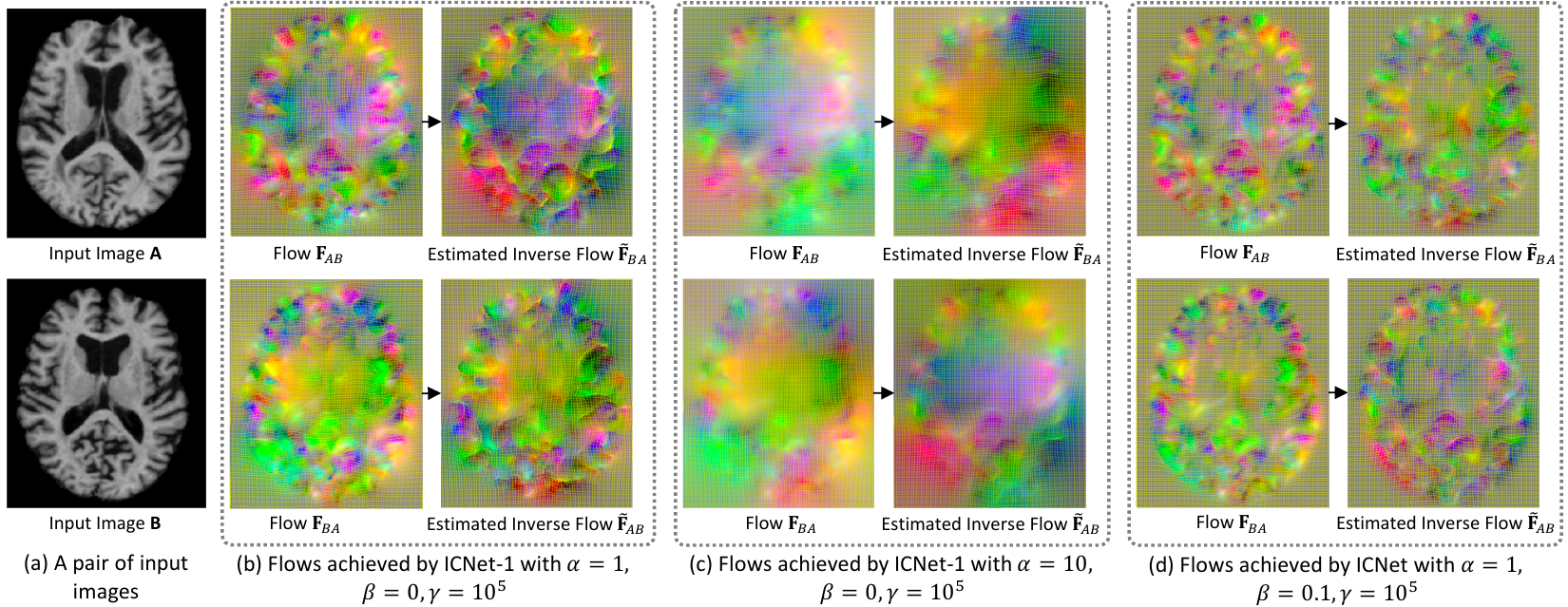}
\caption{Effect of the inverse-consistent constraint. (a) A pair of input images. (b) Flows and estimated inverse flows generated by ICNet-1 with a small weight for the smoothness constraint and without the inverse-consistent constraint (\ie, $\alpha=1$, $\beta=0$ and $\gamma=10^{5}$). (c) Flows and estimated inverse flows yielded by ICNet-1 with a large weight for the smoothness constraint and without the inverse-consistent constraint (\ie, $\alpha=10$, $\beta=0$ and $\gamma=10^{5}$). (d) Flows and estimated inverse flows generated by ICNet with both the smoothness and inverse-consistent constraints (\ie, $\alpha=1$, $\beta=0.1$ and $\gamma=10^{5}$).}
\label{fig_inverseCon}
\end{figure*}

\begin{figure*}[t]
\setlength{\belowdisplayskip}{0pt}
\setlength{\abovedisplayskip}{0pt}
\setlength{\abovecaptionskip}{-1pt}
\centering
\includegraphics[width=1\textwidth]{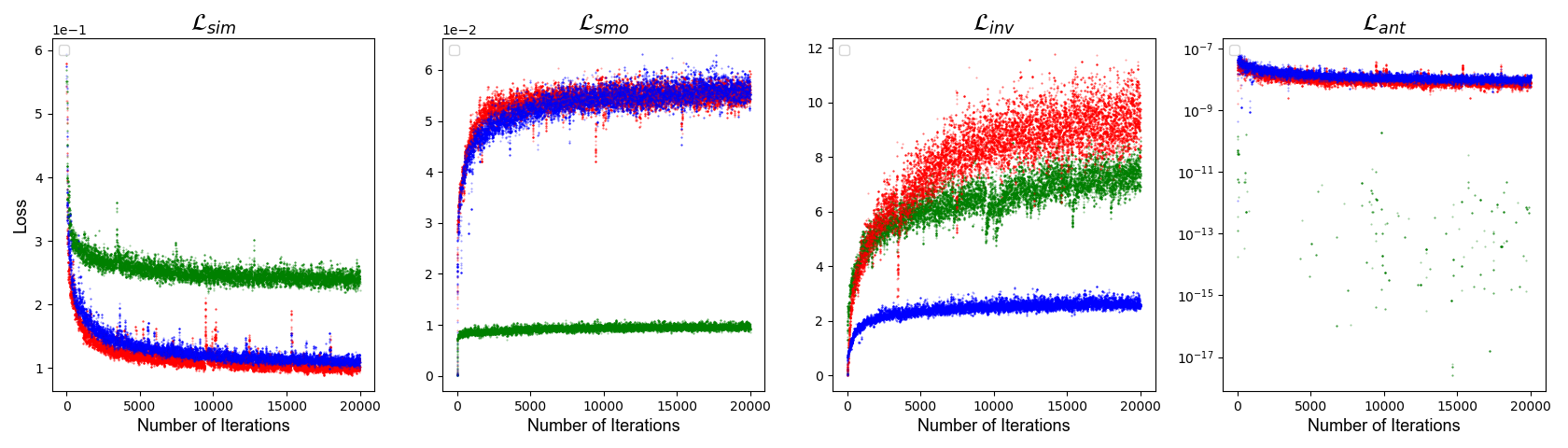}
\caption{Validation losses achieved by ICNet-1 and ICNet using different weights for the smoothing and inverse-consistent constraints, regarding four terms (from left to right) in Eq.~\ref{Eq_loss}. Red line denotes the validation loss of ICNet-1 using a small weight for the smoothness constraint and without the inverse-consistent constraint (\ie, $\alpha=1$, $\beta=0$, and $\gamma=10^5$), green line represents the validation loss of ICNet-1 with a large weight for the smoothness constraint and without the inverse-consistent constraint (\ie, $\alpha=10$, $\beta=0$, and $\gamma=10^5$), and blue line denotes the validation loss of ICNet using a small weight for the smoothness constraint and the inverse-consistent constraint (\ie, $\alpha=1$, $\beta=0$, and $\gamma=10^5$).}
\label{fig_inverseCon_Loss}
\end{figure*}

\subsection{Influence of Inverse-consistent Constraint}
To evaluate the influence of the proposed inverse-consistent constraint in Eq.~\ref{eq_inverseLoss}, we visually illustrate the flows estimated by ICNet with different contributions from the proposed inverse-consistent constraint. Fig.~\ref{fig_inverseCon} shows a pair of input images, as well as the flows and estimated inverse flows achieved by ICNet-1 and ICNet with different parameter settings. Here, we fix the parameter $\gamma=10^5$ for the anti-folding constraint, while the inverse flows are generated by linear interpolation via the proposed inverse network (see Fig.~\ref{fig_framework}~(c)). Results in Fig.~\ref{fig_inverseCon}~(b)-(c) are generated by ICNet-1 without using the inverse-consistent constraint (\ie, $\beta=0$) but having different weights for the smoothness constraint, while those in Fig.~\ref{fig_inverseCon}~(d) are yielded by ICNet with the inverse-consistent constraint.

\begin{figure}[t]
	\setlength{\belowdisplayskip}{-1pt}
	\setlength{\abovedisplayskip}{-1pt}
	\setlength{\abovecaptionskip}{-1pt}
	\centering
	\includegraphics[width=0.49\textwidth]{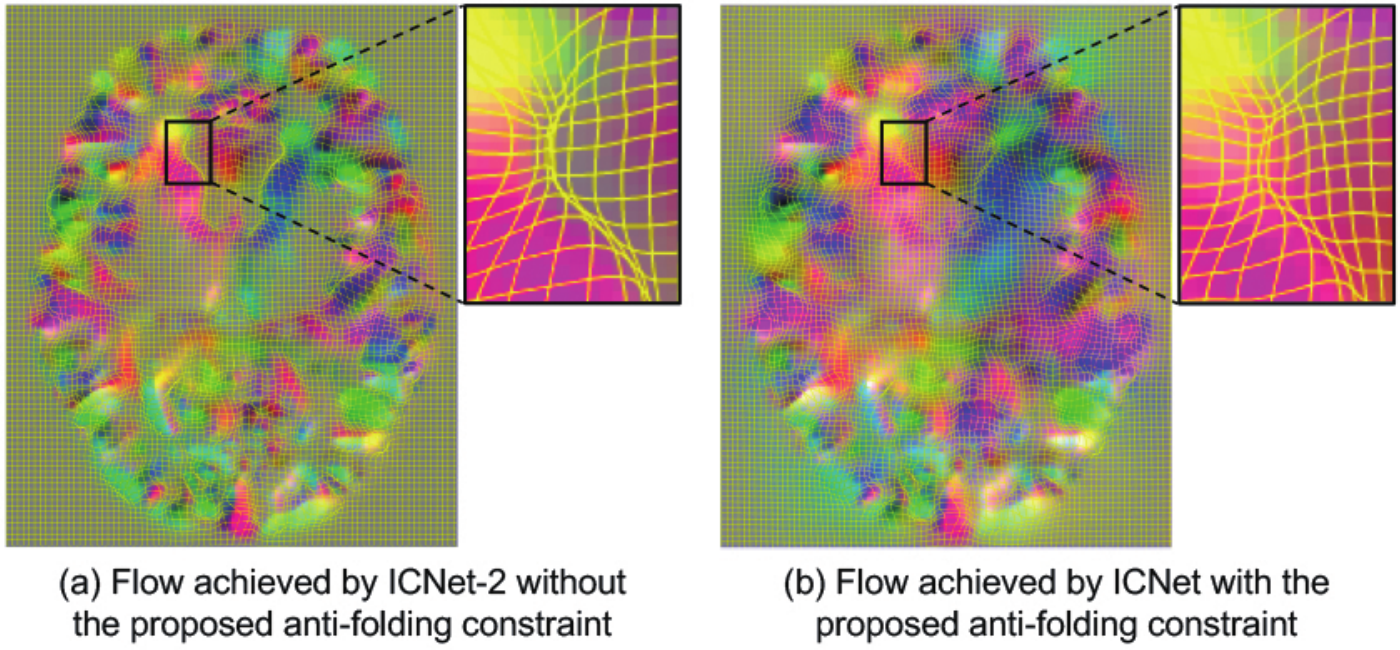}
	\caption{Effect of using the anti-folding constraint, with flows generated by (a) ICNet-2 without the anti-folding constraint (\ie, $\gamma=0$) and (b) ICNet with the anti-folding constraint (\ie, $\gamma=10^5$). Here, $\alpha=1$ and $\beta=0.1$.}
	\label{fig_influenceAntifold}
\end{figure}

\begin{figure*}[htbp]
	\setlength{\belowdisplayskip}{0pt}
	\setlength{\abovedisplayskip}{0pt}
	\setlength{\abovecaptionskip}{-1pt}
	\centering
	\includegraphics[width=0.99\textwidth]{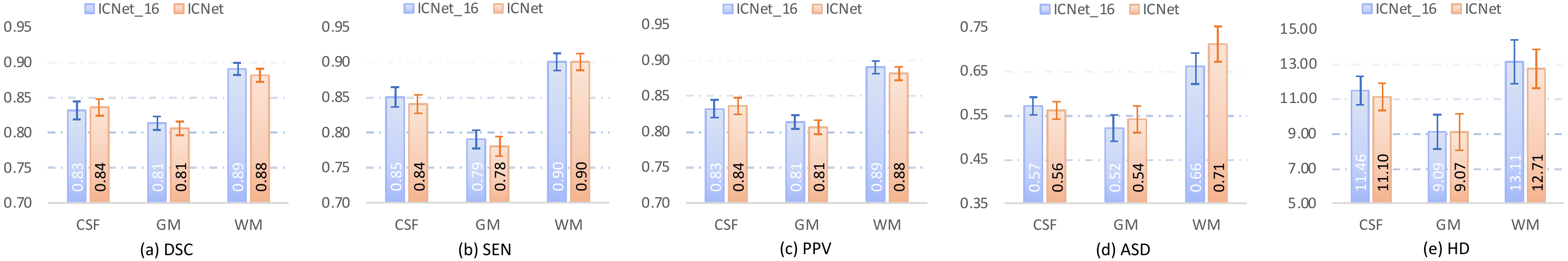}
	\caption{Effect of using different architectures for the proposed method. Here, ICNet has $n=8$ starting filter channels, while ICNet\_16 (a variant of ICNet) includes $n=16$ starting filter channels. (a)-(e) show the tissue segmentation results in terms of five evaluation metrics.}
	\label{fig_influenceChannel}
\end{figure*}

From Fig.~\ref{fig_inverseCon}~(b), we can observe that using a small weight (\ie, $\alpha=1$) for the smoothness term in ICNet-1 cannot generate good results (with foldings in the estimated inverse flow), and also the flow between two images is not inverse-consistent. For instance, the estimated inverse flow $\widetilde{\F}_{AB}$ is not consistent with $\F_{AB}$, while $\widetilde{\F}_{BA}$ looks different from $\F_{BA}$. Fig.~\ref{fig_inverseCon}~(c) suggests that using a large weight (\ie, $\alpha=10$) for the smoothness constraint in ICNet-1 will generate over smooth flow, which may degrade the registration accuracy. Fig.~\ref{fig_inverseCon}~(d) shows that ICNet ($\alpha=1, \beta=0.1$) can generate flows with a reasonable smoothness degree. Also, it can be seen from Fig.~\ref{fig_inverseCon}~(d) that $\widetilde{\F}_{AB}$ is similar to $\F_{AB}$, and $\widetilde{\F}_{BA}$ also looks similar to $\F_{BA}$. This suggests that ICNet can well preserve the inverse-consistent property of the bidirectional flows.

Furthermore, we show the validation loss achieved by ICNet-1 and ICNet with different contributions from the smoothing and inverse-consistent constraints in Fig.~\ref{fig_inverseCon_Loss}. For ICNet-1 with $\beta=0$, the inverse-consistent term in Eq.~\ref{Eq_loss} is not used for network optimization and we only record the corresponding loss here. In Fig.~\ref{fig_inverseCon_Loss}, red line denotes the validation loss of ICNet-1 with $\alpha=1$, green line  represents the loss of ICNet-1 with $\alpha=10$ for the smoothness constraint, and blue line denotes the loss of ICNet with $\alpha=1$ and $\beta=0.1$. As shown in the figure, using a large weight for the smoothing term (green lines) can have relatively small loss $\mathcal{L}_{smo} $, but the $\mathcal{L}_{inv}$ is pretty large. It implies the warped source image is largely different from the target image. Besides, using a small weight for the smoothness regularizer (red lines) can yield relatively large loss $\mathcal{L}_{inv}$, but a good loss $\mathcal{L}_{sim}$ concerning the similarity between the warped source image and the target image. In contrast, the losses of ICNet with $\alpha=1$ and $\beta=0.1$ (blue lines) suggest that ICNet can not only produce inverse-consistent registration, but also keep the warped source image as similar as possible to the target image.

\subsection{Influence of Anti-folding Constraint}
We then study the influence of the proposed anti-folding constraint, by comparing ICNet with ICNet-2 (without using the anti-fold constraint). In this group of experiments, we fix the parameters for the smoothing and inverse-consistent constraints (\ie, $\alpha=1$ and $\beta=0.1$). Fig.~\ref{fig_influenceAntifold} shows the flows generated by ICNet-2 (left) and ICNet (right). It can be observed from Fig.~\ref{fig_influenceAntifold}~(a) that the flow generated by ICNet-2 without using the anti-folding regularizer includes many folding (see the black rectangle) that would degrade the registration accuracy. In contrast, Fig.~\ref{fig_influenceAntifold}~(b) suggests that ICNet using the anti-folding regularizer can effectively avoid the folding in the flow. These results demonstrate the effectiveness of the anti-folding constraint in preventing foldings in the learned transformation.

\subsection{Influence of Network Architecture}
We also investigate the influence of the network architecture on the performance of ICNet, where the number of starting filter channels in FCN (see $n$ in Fig.~\ref{fig_framework}~(b)) is the essential component. In the above-mentioned experiments, we empirically set $n=8$. In this group of experiments, we compare ICNet with its variant denoted as ICNet\_16 with $n=16$, and report the results of tissue segmentation achieved by these two methods in Fig.~\ref{fig_influenceChannel}.

It can be seen from Fig.~\ref{fig_influenceChannel} that ICNet\_16 achieves slightly better results in segmenting the three types of tissues, compared with ICNet. This implies that using more filter channels in the FCN within the proposed ICNet framework helps boost the registration accuracy, thus improving the tissue segmentation performance. Note that ICNet\_16 requires much large memory ($\sim 20GB$) for network training, while the ICNet with $8$ starting filter channels only need $\sim10GB$ memory. Also, the training time of ICNet\_16 will double that of ICNet. Considering the marginal improvement of the registration performance shown in Fig.~\ref{fig_influenceChannel}, we can flexibly choose the number of starting filter channels in practice, based on the memory capacity at hand.

\begin{figure*}[t]
	\setlength{\belowdisplayskip}{0pt}
	\setlength{\abovedisplayskip}{0pt}
	\setlength{\abovecaptionskip}{-1pt}
	\centering
	\includegraphics[width=0.99\textwidth]{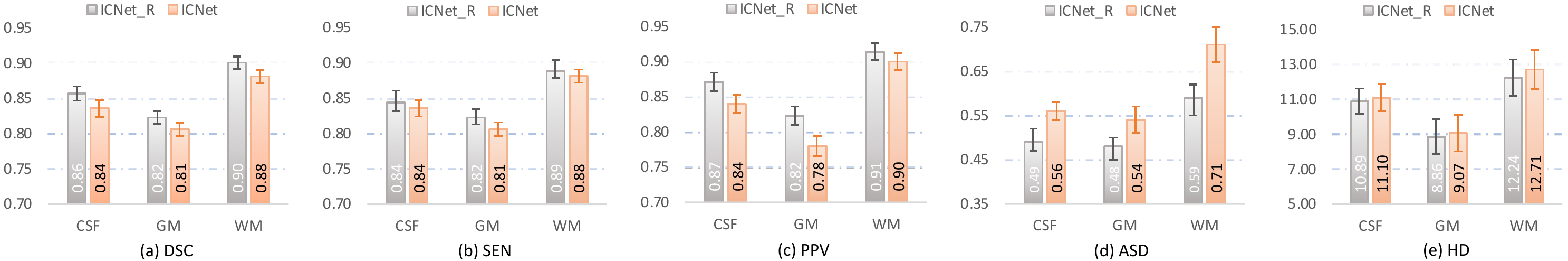}
	\caption{Effect of using refine strategy for the proposed ICNet method. (a)-(e) show the tissue segmentation results in terms of five evaluation metrics. ICNet\_R denotes ICNet with the network refining strategy.}
	\label{fig_influenceRefine}
\end{figure*}

\subsection{Influence of Network Refining Strategy}
The proposed ICNet is unsupervised, without using any ground-truth registration results. Therefore, given a pair of new testing images, we can feed them to ICNet (trained on the training data) to further refine the network, thus adapting the network to the testing images. Here, we denote ICNet with such a network refining process as ICNet\_R. In ICNet\_R, we first optimize the network parameters using the training data, and then refine the network (with the learned parameters as initialization) for the new pair of testing images. In the refining stage, we use a small learning rate (\ie, $1\times10^{-5}$) for optimization, and the number of iteration is empirically set to $100$. After refinement, we can use the newly learned network parameters to produce the final registration results for the testing images. The experimental results on tissue segmentation achieved by ICNet and its refined variant (\ie, ICNet\_R) are shown in Fig.~\ref{fig_influenceRefine}.

It can be seen from Fig.~\ref{fig_influenceRefine} that ICNet\_R consistently outperforms ICNet in segmenting three types of tissues, regarding five evaluation metrics. For instance, the PPV value of ICNet\_R in segmenting CSF is $0.87$, while that of ICNet is only $0.84$. The possible reason could be that the refining strategy makes the network to be better coordinated with the new input images, thus reducing the negative influence of distribution differences between training and test data. It is worth noting that such network refining strategy is a general approach, which can also be applied to improving other unsupervised algorithms for image registration.

\section{Conclusion and Future Work}
\label{S6_conclusion}
In this paper, we propose an inverse-consistent deep network (ICNet) for unsupervised deformable image registration. Specifically, we develop an inverse-consistent constraint to encourage that a pair of images are symmetrically deformed toward one another, and then propose an anti-folding constraint to minimize foldings in the estimated flow. The proposed method is evaluated on registration of T1-weighted brain MR images for tissue segmentation and anatomical landmark detection. Experimental results demonstrate that our ICNet outperforms several state-of-the-art algorithms for deformable image registration.

In the current work, we utilize the mean squared distance (MSD) to measure the similarity between a warped source image and a target image, while many other similarity measures (such as correlation or mutual information) can be employed in the proposed deep-learning framework. Besides, only a simple network refining strategy is proposed in this work to handle the challenge of data heterogeneity (\eg, in image contrast, resolution, and noise level), while more advanced data adaptation algorithms~\cite{gopalan2014unsupervised,jhuo2012robust} could be used to further boost the registration performance.

\section*{Appendix}

As mentioned in the main text, besides using the conventional smoothness constraint~\cite{de2017end}, we also develop an anti-folding constraint to further prevent the folding in the learned discrete displacement field (\ie, flow). Note that each element in a flow is a $3$-dimensional vector (corresponding to three axes, \ie, $x,y,z$), indicating the displacement of a particular voxel from its original location to a new location. We now explain the details of this anti-folding constraint defined in Eq.~$5$ in the main text.

\begin{figure}[htbp]
	\setlength{\belowdisplayskip}{0pt}
	\setlength{\abovedisplayskip}{0pt}
	\setlength{\abovecaptionskip}{-2pt}
	\centering
	\includegraphics[width=0.38\textwidth]{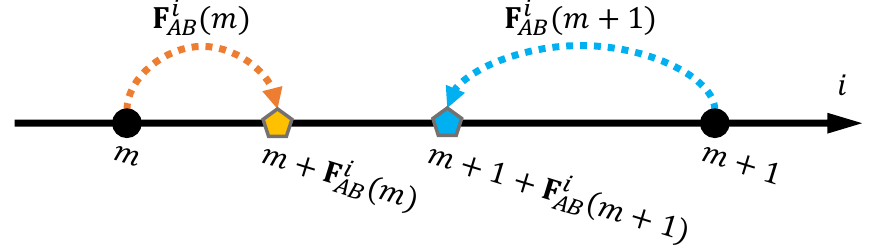}
	\caption{Illustration of the proposed anti-folding constraint, with an index function to penalize the gradient along the $i$-th axis to avoiding folding at the location $m$. Here, $m+1$ denotes the nearest neighbor of $m$. To avoiding folding at the location of $m$, after the displacement $\F^i_{AB}(m)$ for $m$ and $\F^i_{AB}(m+1)$ for $m+1$, the new locations of these two points should follow that $m+\F^i_{AB}(m) < m+1+\F^i_{AB}(m+1)$.}
	\label{fig_gradient}
\end{figure}

We denote the to-be-estimated flow from an image $\A$ to an image $\B$ as $\F_{AB}$, and $\F_{AB}$ is defined in the $\A$ space. For simplicity, we denote the $1$-dimensional displacement in $\F_{AB}$ along the $i$-th axis as $\F^i_{AB}$. As shown in Fig.~\ref{fig_gradient}, for the point $m$ and its nearest neighbor point $m+1$ along the $i$-th axis in the space of $\A$, we denote their displacements as ${\F^i_{AB}\left(m\right)}$ and ${\F^i_{AB}\left(m+1\right)}$, respectively.

To avoid folding at the location of $m$, it is required that the new locations of these two points (after displacement via $\F^i_{AB}$) should follow
\begin{equation}
\small
\begin{aligned}
& m+{\F^i_{AB}\left(m\right)} < m+1+{\F^i_{AB}\left(m+1\right)}\\
\Rightarrow\,&{\F^i_{AB}\left(m+1\right)} - {\F^i_{AB}\left(m\right)} +1 > 0
\end{aligned}
\label{location}
\end{equation}
where $m+{\F^i_{AB}\left(m\right)}$ is the new location of the point $m$, and $(m+1)+{\F^i_{AB}\left(m+1\right)}$ denotes the new location of the point $m+1$. In discrete problems, the gradient of $\F^i_{AB}$ at the location of $m$ along the $i$-th axis is typically defined as
\begin{equation}
\small
\begin{aligned}
 \nabla{\F^i_{AB}\left(m\right)} =&\, \frac { {\F^i_{AB}}(m+1) - {\F^i_{AB}}(m)}{(m+1)-m} \\
 =  &\,{\F^i_{AB}}(m+1) - {\F^i_{AB}}(m)
\end{aligned}
\label{gradient}
\end{equation}

Combining Eq.~\ref{gradient} to Eq.~\ref{location}, we can get the following
\begin{equation}
\small
\begin{aligned}
\nabla{\F^i_{AB}\left(m\right)} +1 > 0
\end{aligned}
\label{noFolding}
\end{equation}
which indicates that, if $\nabla{\F^i_{AB}\left(m\right)} +1 > 0$, there is no folding at the location of $m$. In contrast, if $\nabla{\F^i_{AB}\left(m\right)} +1 \leq 0$, there is folding at the location of $m$.

Accordingly, to avoid folding in the learned flow, we propose an anti-folding constraint by penalizing the gradient of the flow, at locations that violate the rule in Eq.~\ref{location}. Let $Q=\nabla{\F^i_{AB}\left(m\right)}+1$, and we denote an index function $\delta{\left(Q\right)}$, where $\delta{\left(Q \right)} = |Q|$ if $Q\le0$, and $\delta{\left(Q\right)}=0$, otherwise. Specifically, in the proposed anti-fold constraint (see Eq.~$5$ in the main text), if there is folding in the location of $m$ along the $i$-th axis (\ie, $\nabla{\F^i_{AB}\left(m\right)}+1\le0$), we enforce the penalty $|\nabla{\F^i_{AB}\left(m\right)}+1|$ on the gradient at this location, requiring this gradient to be small. In contrast, if there is no folding at the location of $m$ along the $i$-th axis (\ie, $\nabla{\F^i_{AB}\left(m\right)}+1>0$), we do not penalize the gradient of the estimated flow at this location.

\end{document}